\documentclass[journal,10pt]{IEEEtran}
\usepackage[pdftex]{graphicx}
\usepackage{multirow}
\usepackage{amsmath}
\usepackage{mathtools}
\usepackage{optidef}
\usepackage{epstopdf}
 \usepackage[ruled,vlined]{algorithm2e}
  \usepackage[table]{xcolor}
\usepackage{array}
\newcolumntype{P}[1]{>{\centering\arraybackslash}p{#1}}
\newcolumntype{M}[1]{>{\centering\arraybackslash}m{#1}}
\usepackage{subcaption}
\usepackage{authblk}
\usepackage{cite}
\usepackage{ctable} % for \specialrule command
\usepackage{booktabs}
\usepackage{cleveref}
\usepackage{latexsym}
\usepackage[justification=centering]{caption}
\usepackage{amsthm}
\usepackage{balance}
\usepackage{relsize}
\usepackage[nodisplayskipstretch]{setspace}
\usepackage{color}
\usepackage{amssymb}
\usepackage{eucal}
\usepackage{url}
\usepackage{booktabs,subcaption,amsfonts,dcolumn}

\usepackage{float}
\usepackage{scalefnt}
\usepackage{geometry}

\usepackage[nodisplayskipstretch]{setspace}
\usepackage{color}  
\definecolor{Mypink}{RGB}{255,0,255}
\definecolor{Myorange}{RGB}{255,102,0}
\definecolor{Mygreen}{RGB}{0,153,0}
\definecolor{Myblue}{RGB}{0,0,255}
\newcolumntype{d}[1]{D..{#1}}
\usepackage{lipsum}

\DeclareMathAlphabet\mathbfcal{OMS}{cmsy}{b}{n}
\usepackage{breqn}

\begin{document}

\title{Towards Federated Learning-Enabled Visible Light Communication in 6G Systems}
\renewcommand\Authfont{\fontsize{12}{14.4}\selectfont}

\author{Shimaa Naser,~\IEEEmembership{Student Member,~IEEE,} Lina Bariah,~\IEEEmembership{Senior~Member, IEEE,}
Sami Muhaidat,~\IEEEmembership{Senior Member,~IEEE,}  Mahmoud Al-Qutayri, \IEEEmembership{Senior Member, IEEE}, Ernesto Damiani,~\IEEEmembership{Senior Member,~IEEE,}   
M\'erouane Debbah,~\IEEEmembership{Fellow,~IEEE}, and
Paschalis C. Sofotasios,~\IEEEmembership{Senior Member,~IEEE}

\thanks{This work was supported  by Khalifa University
under Grant KU/FSU-8474000122 and Grant KU/RC1-C2PS-T2/8474000137. }

\thanks{S. Naser, L. Bariah, and E. Damiani are with the Center for Cyber-Physical Systems, Department of Electrical Engineering and Computer Science, Khalifa University, Abu Dhabi 127788, UAE, (e-mails: $ \rm \{100049402; lina.bariah; ernesto.damiani\}$@ku.ac.ae).}

\thanks{S. Muhaidat is with the Center for Cyber-Physical Systems, Department of Electrical Engineering and Computer Science, Khalifa University, Abu Dhabi 127788, UAE, and also with the Department of Systems and Computer Engineering, Carleton University, Ottawa, ON K1S 5B6, Canada, (e-mail: muhaidat@ieee.org).}

\thanks{M. Al-Qutayri is with the Systems-on-Chip (SoC) Center, Department of Electrical Engineering and Computer Science, Khalifa University, Abu Dhabi 127788, UAE, (e-mail: mahmoud.alqutayri@ku.ac.ae).}

\thanks{M. Debbah is with the Technology Innovation Institute, 9639 Masdar City,
Abu Dhabi, UAE and also
with Centrale Supelec, University Paris-Saclay, 91192 Gif-sur-Yvette, France, (email: merouane.debbah@tii.ae).}

\thanks{P. C. Sofotasios is with the Center for Cyber-Physical Systems, Department of Electrical Engineering and Computer Science, Khalifa University, Abu Dhabi 127788,  UAE, and also with the Department of Electrical Engineering, Tampere University, Tampere 33014, Finland, (e-mail: p.sofotasios@ieee.org).}
}\maketitle

\begin{abstract}
\textcolor{black}{
Visible light communication (VLC) technology was introduced as a key enabler for the next generation of wireless networks, mainly thanks to its simple and low-cost implementation. However, several challenges prohibit the realization of the full potentials of VLC, namely, limited modulation bandwidth, ambient light interference, optical diffuse reflection effects, devices non-linearity, and random receiver orientation. On the contrary, centralized machine learning (ML) techniques have demonstrated a significant  potential in handling different challenges relating to wireless communication systems. Specifically, it was shown that ML algorithms exhibit superior capabilities in handling complicated network tasks, such as channel equalization, estimation and modeling, resources allocation, and opportunistic spectrum access control, to name   a few. Nevertheless, concerns pertaining to privacy and communication overhead when sharing raw data of the involved clients with a server constitute major bottlenecks in the implementation of centralized ML techniques. This has motivated the emergence of a new distributed ML paradigm, namely federated learning (FL), which  can reduce the cost associated with transferring raw data, and preserve privacy by training ML models locally and collaboratively at the clients' side. Hence, it becomes evident that integrating FL into VLC networks can provide ubiquitous and reliable implementation of VLC systems.
With this motivation, this is the first in-depth review in the literature on the application of FL in VLC networks. To that end, besides the different architectures and related characteristics of FL, we   provide a  thorough overview on the main design aspects of FL based VLC systems. 
Finally, we also highlight some potential future research directions of FL that are envisioned to substantially  enhance  the performance and robustness of VLC systems.
}
\end{abstract}
%\begin{IEEEkeywords}
%\end{IEEEkeywords}

\section{Introduction}
\IEEEPARstart{T}he recent advancements in indoor lighting systems, accompanied with the revolutionary solid-state progression in light emitting diodes (LEDs), have motivated the emergence of the   visible light communication (VLC) concept. In particular, the ability of LEDs to switch rapidly between different light intensity levels enables short range data   transmission    without affecting their illumination capability, rendering VLC   a cost-effective easy-to-implement technology. VLC systems have enabled a swarm of wireless applications, such as indoor navigation, healthcare, underwater communication, positioning systems, and vehicular communications. The key driver underlying the emergence of such applications are the promising features of VLC systems, such as enhanced capacity, inherent secure communication and ultra-low end-to-end latency. \\ 
\indent 
Likewise,  machine learning (ML) is a sub-field of artificial intelligence (AI), which has been   recently identified as an appealing data-driven solution for optical wireless networks \cite{musumeci2018}. In  centralized ML algorithms, mobile nodes share their data, which are then uploaded, processed, trained, and aggregated in cloud-based servers. Nevertheless, the drawbacks of such cloud-centric algorithms are threefold. Firstly, data privacy is compromised in cloud-based ML because participating devices are requested to send their private data to centralized servers, exposing these data to potential eavesdroppers. Secondly, centralized ML suffers from long propagation delay, rendering it unsuitable for real-time applications. Thirdly, data transmission yields increased network overhead, which renders the implementation of  centralized ML in resources-constrained internet-of-things (IoT) devices challenging. This ultimately calls for a fundamental transition from conventional centralized algorithms into novel paradigms, in which networks can be trained in a distributed manner. 
\begin{table*}[t]
\caption{Recent surveys on FL.}
\label{TableI}
\centering
\footnotesize
\begin{tabular}{ |>{\color{black}} M{1.5cm}| >{\color{black}} M{10cm} |>{\color{black}}M{4cm}|}
\hline 
{\textbf{Reference}}&{\textbf{Main focus }} & {\textbf{Application}} \\
\hline 
\hline

{\cite{Wahab2021}}&{\begin{itemize}
    \item FL concept, technologies and learning approaches.
    \item Introducing a three-level classification of FL approaches.
    \item Finally, highlighting on some applications and future works of FL.
\end{itemize} }&{Communication and Networking} \\ \hline
{\cite{kairouz2021}}&{Recent advances, challenges, and open research problems in FL.}&{-}  \\ \hline
{\cite{Qinbin2021}} & {Definition of FL and its categorization based on six-aspects. }&{Mobile services, Healthcare, and Finance.}  \\ \hline
{\cite{yang2019}}&{Categorization of different FL settings, namely vertical FL, horizontal FL, and federated transfer learning.  }&{Smart retail, Smart
Healthcare, Financial services, and Mobile content predictions.} \\ \hline
{\cite{Lim2020}}&{Background, fundamentals, and challenges of FL in mobile edge networks}&{Networking}  \\ \hline
{\cite{Li_2020}} & {FL, its characteristics, challenges, and future  directions in  massive  networks.
}&{Networking}  \\ \hline
{\cite{niknam2020}} & {FL concept, its  applications, key technical challenges and open research problems in 5G networks.
}&{Communication}  \\ \hline
{\cite{Aledhari2020}} & {FL  and  its  enabling  software  and hardware platforms, protocols, real-life applications and use-cases. } &{Various industrial purposes, including
healthcare.} \\ \hline
 \cite{lyu2020} & {Threats models and major attacks in FL.} &{Business}\\   \hline
 \cite{Posner2021}& {\begin{itemize}
     \item Federated vehicular networks, their high-level architecture and enabling technologies.
     \item Introducing blockchain-based systems to mitigate any malicious behaviour.
     \item Discussing possible future research directions and open problems.  
 \end{itemize} }&{Networking}  \\ \hline
\end{tabular}
\end{table*}

In this respect, \textit{Federated Learning} (FL) was recently considered   an efficient tool to train wireless networks, without leaking private information or consuming network resources. Specifically, the enhanced on-board computational and storage capabilities of mobile devices with the local datasets are leveraged to enable decentralized local  training. \textcolor{black}{However, most reported investigations focused on integrating FL into radio frequency (RF) systems; hence FL integration in VLC networks has not received significant attention. Yet, adopting FL in VLC is particularly appealing in accommodating the ever-growing demands of data-hungry, privacy-sensitive applications, such as extended reality (XR), flying vehicles, telemedicine, connected autonomous systems, and industrial internet of things, to name a few. The key features of VLC, such as the inherent security, high transmission data rate, energy efficiency, and high spatial reuse are key drivers for the integration of FL into VLC systems. Through offloading data traffic from the congested RF bands, VLC can provide secure, reliable, and fast global model evaluation for FL process. In fact, the relationship between FL and VLC is bidirectional, in the sense that FL offers many benefits towards boosting the performance of VLC systems through handling different complicated tasks, such as resources management, network control, interference
alignment and user grouping.} With this motivation, this article  presents a thorough overview of the integration of FL into VLC systems along with interesting and useful theoretical and practical insights.   
\subsection{Related Work}
Inspired by the promising advantages of FL for communication and networking, significant research efforts have been devoted to explore FL in terms of architectures, challenges, design aspects, and applications. Particularly,  \cite{Wahab2021} presented a comprehensive survey that highlights the fundamentals, applications, enabling technologies, and learning mechanisms of FL. On the contrary,  \cite{kairouz2021}  discussed  open research problems and challenges associated with FL, such as communication efficiency, data privacy, data heterogeneity and model aggregation. From a different perspective,  \cite{Qinbin2021} presented the FL taxonomy, with emphasis on the main FL components, including data distribution, ML model, privacy mechanism, communication architecture, scale of federation, and motivation of federation. In \cite{yang2019}, FL was explored from a security and privacy perspective, while \cite{Lim2020} discussed FL in mobile edge computing. In the same context,   \cite{Li_2020} considered the characteristics, challenges, and future directions of FL in massive-scale networks, while \cite{niknam2020} addressed the implementation and applications of FL in fifth generation (5G) networks. A comprehensive study of FL and its enabling software and hardware platforms, protocols, real-life applications and use-cases was carried-out in \cite{Aledhari2020}. Moreover, the key techniques and  fundamental assumptions adopted by various attacks in FL were explained in \cite{lyu2020}. Then, the authors discussed future research directions towards more robust privacy preservation in FL. \textcolor{black}{Finally,  the enabling technologies
of federated vehicular networks (FVN) were outlined in \cite{Posner2021}, in which a high-level architecture of FVN was discussed.} 
For convenience, the related reported contributions are summarized in Table \ref{TableI}.\\ 
\indent
The aforementioned contributions  considered the implementation of FL in RF scenarios. Yet, the interplay between FL and VLC systems is barely addressed in the open literature. Accordingly,   this article provides an overview on the implementation of FL in VLC systems, and sheds light on  the associated design and deployment aspects. Furthermore, it  constructs a road-map towards  open research directions that require thorough investigation. 
\begin{figure}[t]
        \includegraphics
       [width=1\linewidth] {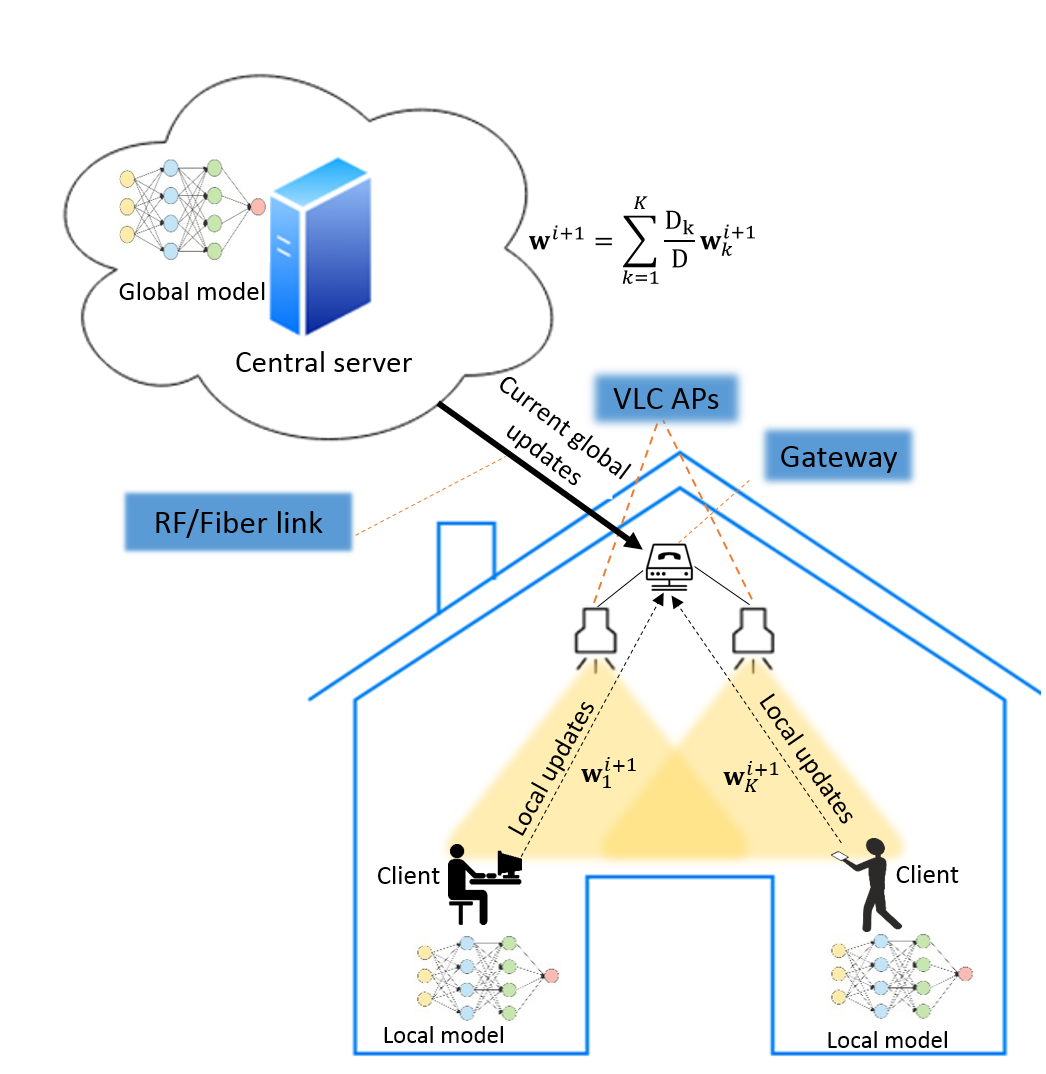}
    \caption{{System architecture of FL in VLC.}}
    \label{model}
\end{figure} 
\section{Federated learning in vlc}
%In this section, we present our vision on the integration of FL in VLC systems, with pointing out the commonly adopted global model aggregation technique. 
\subsection{Fundamentals}

A typical VLC system that employs FL is illustrated in Fig. \ref{model}. The appealing features of VLC, such as inherent security, high data rate transmission, energy efficiency and high spatial reuse are the main drivers for utilizing VLC with FL, in order to provide secure, accurate, and fast global model evaluation.

In the context of VLC, multiple LEDs connected through RF or optical fiber links to a gateway and then to a server represent an interface between the cloud-based server and the participating clients. Specifically,  the LEDs are exploited at the downlink communication to assist with global model transmission. Accordingly, the participating clients, which are equipped with photo-detectors (PDs) to receive the global model, are determined based on the field of view (FoV) of the LEDs. Therefore, clients existing in the LEDs' coverage area can only take part of the learning process. Also,  local model updates are communicated with a gateway, connected to the cloud-based server, through uplink RF or infrared links. This is attributed to the clients limited energy and the undesired radiance from the clients devices. On the contrary, VLC can be leveraged to offload model update traffic in the downlink from the overcrowded RF spectrum to the visible light band, allowing enhanced allocation of the bandwidth resources in the uplink communication. 

The centralized cloud-based server in such scenarios is responsible of fulfilling the following tasks: i) initialize the global model evaluation process for a particular learning task; ii) select the participating clients based on different metrics, including the LEDs' coverage area, clients mobility, clients' receivers orientation, etc; iii) coordinate the learning process and model aggregation. 
Hence, $K$ clients, out of a set comprising $N$ nodes, are selected to receive the initial global model parameters $\mathbf{{w}}^o$, aiming to engage them in the learning process. The $k^{th}$ client will utilize its dataset $\mathfrak{D}_k$, which is stored locally, for training. Each dataset is assumed to be composed of $D_k$ input-output pair vectors $(\mathbf{x}_k,\mathbf{y}_k)$. Assuming stochastic gradient descent at the $i^{th}$ communication round, the $k^{th}$ client calculates the gradient of the loss function.
%, namely
%\begin{equation}
 %   F_k(\mathbf{w}^i)= \frac{1}{D_k} \sum_{j\in \mathfrak{D}_k} f_{jk}(\mathbf{w}^i)  
%\end{equation}
%with $\mathbf{x}_k = [x_{k1}, ...x_{kD_k}]$ and $\mathbf{y}_k = [y_{k1}, ...y_{kD_k}]$, where $x_{kj}$ and $y_{kj}$ are the $j^{th}$ input-output vectors of the FL at the $k^{th}$ client
%where $f_{jk}(\mathbf{w}^i)$ is the loss function of the $j^{th}$ input-output sample, which could be a linear regression, neural network, or logistic regression. 
%with $f_{jk}(\mathbf{w}^i)= \frac{1}{2}( x_{kj}^{\rm T} \mathbf{w}^i- y_{kj})^2$, or non-convex function in the case of a neural networks. 
\subsection{Model Aggregation}
\begin{figure}[t]
\centering
        \includegraphics [width=400 pt] {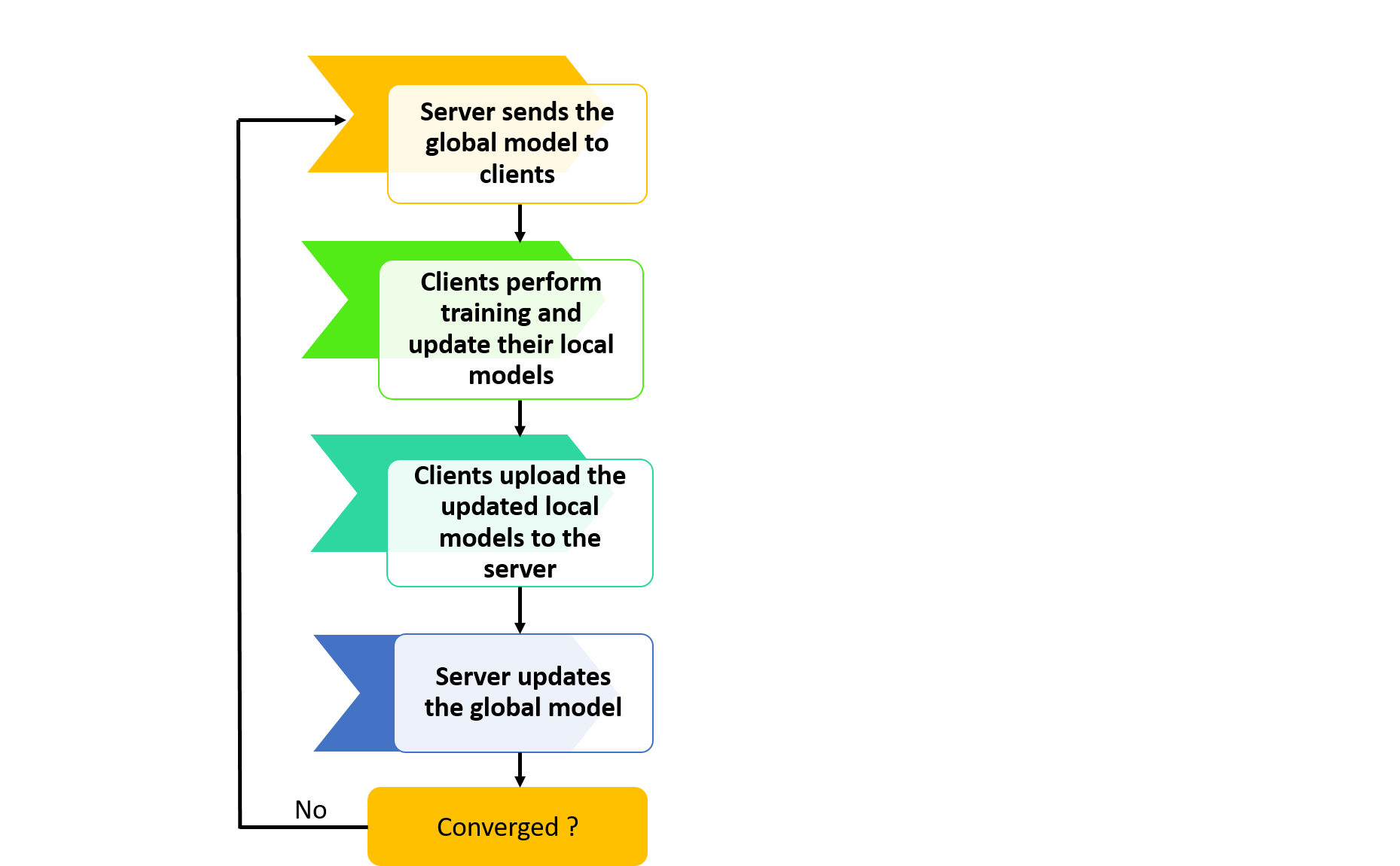}
    \caption{Learning steps in \textcolor{black}{FL}.}
\label{Learningmechanism}
\end{figure}

After receiving all local gradient updates from participating clients in the $i^{th}$ communication round, the centralized server performs aggregation in order to compute the global model parameters. An aggregation model, referred to as \textit{federated averaging} (\textit{FedAvg}), is used for this in which all local parameters are combined using model averaging. %\cite{mcmahan2017}.
%To calculate the new global model parameters, the centralized server performs the following operation  
%\begin{equation}
 %   F(\mathbf{w}^{i+1})= \frac{1}{D} \sum_{k=1}^K  { D_k F_k(\mathbf{w}^i_k) }
%\end{equation}
%\begin{equation}
%\label{eq:modelAgg}
  %  \mathbf{w}^{i+1}=\mathbf{w}^{i}-\eta \frac{1}{D} \sum_{k=1}^K  { D_k \Delta F_k(\mathbf{w}^i) }
%\end{equation}
%where $D$ is the total data size of participating clients, $\eta$ is the learning rate, and $\Delta$ is the derivative operation. 
Thus, the server updates the global model parameters based on the weighted average of the attained local parameters.  
%\begin{equation}
 %  \mathbf{w}^{i+1}=\mathbf{w}^{i} - \eta \Delta F(\mathbf{w}^{i})=  \frac{\sum_{k=1}^K D_k \mathbf{w}^i_k }{D}
%\end{equation}centralized lea
Following this, the server shares the updated global model parameters with the clients in the next iteration  to enhance the accuracy of the global model. Notably,  weights exchange is performed over multiple rounds until a certain model accuracy level  is satisfied. For convenience, the typical learning steps in FL are summarized in Fig.\ref{Learningmechanism}. It should be noted that several variants of aggregation models have been proposed to enhance the performance of \textit{FedAvg} scheme, such as \textit{FedProx, FedPAQ, Turbo-Aggregate, FedMA,} and \textit{HierFAVG}, \textcolor{black}{which are summarized in Table \ref{TableII}. } 

\textcolor{black}{Based on the distribution characteristics of the datasets, FL can be categorized into horizontal FL, vertical FL, and federated transfer learning. Additionally, different versions of the centralized FL architecture were proposed in order to  overcome clients failure, system
scalability, and communication efficiency challenges. These include, hierarchical, regional
and decentralized architectures \cite{zhang2020}. Different FL categories and architectures are summarized in Table \ref{TableII}.}

\section{Design aspect of fl in vlc}
%In this section, we explore some design aspects and challenges associate with the integration of FL with VLC systems, with highlighting potential solution to overcome these challenges.
\subsection{Client Selection and Scheduling}
Client selection and scheduling constitute an important factor in the implementation of FL in VLC because of their effect on the accuracy and convergence time of the training process. In this regard, the need for developing efficient client selection and scheduling schemes stems from the heterogeneity of clients datasets, devices diverse computational capabilities, available resources, and wireless channel conditions. 

Thus, a random selection of clients to participate in the FL process \textcolor{black}{without considering their diverse computational, communication, and storage capabilities}, could reduce its efficiency. For example, selecting a client with limited computational capabilities or severe channel conditions will require additional time to compute  its updated local model parameters and  send them to the server, whilst it will drain the client resources.  Accordingly, this will result in a delayed global model aggregation procedure, needed to accomplish the scheduled training process. \textcolor{black}{ This challenge is usually referred to as \textit{system heterogeneity}}. Hence, a key factor to improve the training convergence time and achieve a high-performance training relies on how to properly select the participating clients and assign the training tasks among them. 

Efficient clients selection and scheduling should be performed while considering  numerous aspects. Typical VLC indoor environment is usually deployed with multiple LEDs, each with limited coverage area. Therefore, in order to provide ubiquitous coverage each LED acts as a  VLC access point (AP) that handles multiple clients located in its coverage area. Consequently, clients and AP association is the first step towards realizing efficient client selection and scheduling, and subsequently successful implementation of FL in indoor VLC environments. It is also recalled that in order to achieve acceptable performance in FL, a considerable number of clients should participate in the local training process. Consequently,  this will lead to an increased communication overhead, due to the limited bandwidth of   uplink and downlink channels, as well as   constrained energy resources. Hence, developing effective resource management schemes for   uplink and downlink links is essential in minimizing  resources consumption, while maximizing global model accuracy. Also, most of the available resource management techniques rely on formulating optimization problems that are handled by heuristic or reinforcement learning tools. As most of these techniques are developed for RF communications, their extensions to VLC systems need to  be revisited. 

From a different angle, a larger number of clients does not readily imply faster model convergence, due to the increased heterogeneity of the data among clients and the waiting time, which may cause additional delay. Moreover, reliability of the participating clients is another issue related to increasing the number of the clients. Hence, the optimum number of participating clients should be maximized while taking into consideration these concerns. 
In this context, \textit{straggler clients}, which are dropped from the learning process due to their low battery levels, the absence of \textcolor{black}{line-of-sight} (LoS) links, or connectivity issues, is a common problem in FL that yields wasted resources of both the server and other clients. Several research contributions  shed  light on these issues, proposing numerous techniques to overcome the straggler clients problem, including, redundancy and asynchronicity. Finally, given that global parameters in the downlink are carried over the light intensity, maintaining the communication while lights are off by switching to RF guarantees successful implementation of FL in VLC. Specifically, a switch to conventional RF clients scheduling schemes should be implemented in the off-light mode. 
%To further reduce the communication overhead, either reducing the total number of communication
%iterations, or reducing the size of the exchanged models in each iteration were investigated in the literature \cite{Li_2020}. 
%%%%%%%%%%%%%%
\begin{table*}
\footnotesize
\begin{subtable}[c]{0.5\textwidth}
\centering
\begin{tabular}{|>{\color{black}} M{2cm}| >{\color{black}} M{11cm} |>{\color{black}}M{3cm}|}
\hline
\rowcolor{lightgray}{\textbf{Aggregation method}}&{\textbf{Main idea }} &{\textbf{Tackled challenge}} \\
\hline \hline
{FedAvg}&{ Weights of multiple local
models are averaged by the server to calculate the new global model updates.} &{Data heterogeneity}\\ \hline
{FedProx}&{ Generalizes FedAvg
by allowing variable amounts of work to be performed
locally across devices based on their available system resources also a proximal term is used to  stabilize the method.}&{Data
heterogeneity} \\ \hline
{FedPAQ}&{Clients perform multiple local updates on
the model before sharing the weights with the server.}&{Communication efficiency} \\ \hline
{Turbo-Aggregate}&{Based on a multi-group training, i.e., users are divided into several groups where the model updates are shared between different groups in a circular manner. Also, a secret sharing and novel coding techniques are used.  }&{Communication efficiency and security} \\ \hline
{FedMA}&{It accounts for permutation invariance of
the neurons and enables global model size adaptation. }&{Data heterogeneity } \\ \hline
{HierFAVG}&{Allows multiple servers to perform partial model aggregation.}&{Communication efficiency} \\ \hline \hline
\end{tabular}
%\subcaption{subtable no. 1}
\end{subtable}

 \begin{subtable}[c]{0.5\textwidth}
\centering
\begin{tabular}{|>{\color{black}} M{2cm}| >{\color{black}} M{12cm} |>{\color{black}}M{2cm}|}

\rowcolor{lightgray}{\textbf{Category}}&{\textbf{Main features }} &{\textbf{Focus}} \\
\hline \hline
{Horizontal FL}&{Datasets of different clients share the same feature space but have different sample space. } &{Security}\\ \hline
{Vertical FL}&{Datasets of different clients share the same sample space but differ in feature space.}&{Privacy} \\ \hline
{Federated Transfer
Learning }&{Datasets of different clients differ not only in the sample space but also in feature space.}&{Reduce accuracy loss} \\ \hline \hline
\end{tabular}
%\subcaption{subtable no. 1}
\end{subtable}

\begin{subtable}[c]{0.5\textwidth}

\begin{tabular}{ |>{\color{black}} M{2cm}| >{\color{black}} M{6cm} |>{\color{black}}M{4cm}|>{\color{black}}M{3.6cm}|}
\hline 
\rowcolor{lightgray}{\textbf{Architecture}}&{\textbf{Main features }} &{\textbf{Advantages}} &{\textbf{Drawbacks}}  \\
\hline \hline
{Centralized}&{Single central server  is
responsible of the communication with the local clients, aggregating
local models updates, and sharing the global model.} &{\begin{itemize}
    \item {Model transmission is smooth.}
    \item{System can be easily modified to suit customized tasks.}
    \item{Any local client can be easily detached from the learning process.}
\end{itemize}
}&{\begin{itemize}
    \item Scalability limitations.
    \item Increased communication overhead.
    \item Single-point of failure.
\end{itemize}  }\\ \hline
{Hierarchical}&{ Multiple regional coordinated nodes are employed to handle edge clusters. The role of the central server is limited to sending global model updates. }&{\begin{itemize}
    \item Reduces communication overhead compared to the centralized FL approach.
\end{itemize}}&{\begin{itemize}
    \item Increased management cost.
    \item The need for more aggregation servers.
    \item Single-point of failure.
\end{itemize} } \\ \hline
{Regional }&{Similar setup as hierarchical architecture, but without considering the central server. Models are aggregated and
exchanged via regional aggregation nodes assigned to each clients cluster. }&{\begin{itemize}
    \item Computationally efficient.
\item  Overcomes single-point of failure. 
\end{itemize}}&{\begin{itemize}
    \item Increased hardware cost and server configuration management.
\end{itemize}} \\ \hline
{Decentralized }&{
Consists of edge nodes only, and the
aggregation process is moved to the local clients side.}&{\begin{itemize}
    \item Adapts easily to environmental changes.
    \item Reduces performance bottleneck.  
\end{itemize}}&{\begin{itemize}{ 
    \item Lack of coordination between clients. Therefore, it is difficult to handle collective tasks and global knowledge. }
\end{itemize}}  \\ \hline
\end{tabular}
%\subcaption{subtable no. 1}
\end{subtable}
\caption{Summary of FL aggregation methods, and different FL categories and architectures. }
\label{TableII}
\end{table*}
\subsection{Joint Communication and Learning} 
The communication process in FL for VLC is carried out over two different wireless media. Specifically, the first one is the downlink communication which is realized through optical signals for sharing the global model parameters after aggregating them at the server. 
The uplink is   realized over RF or infrared signals for uploading the updated local models to the server. Indeed, communication over wireless media is usually unreliable due to the effect of different impairments such as noise, shadowing, fading, and path loss. In addition to that, VLC introduces additional impairments, including, ambient light interference and random receiver orientation. Hence, the accuracy of the model and the convergence time of FL is highly dependent upon the channel impairments, that may introduce significant training errors.
 
Therefore, in order to ensure a realistic and accurate implementation of FL in VLC systems, the effect of transmission errors in the uplink and the downlink needs to  be addressed. To this end, different error detection codes such as, parity checking, cyclic redundancy check, or longitudinal redundancy check can be utilized, aiming to determine if the global and local models are erroneously received. Subsequently, the server will discard the invalid local model updates, and aggregate the error-free local model updates only. Similarly, clients who experience a degraded optical signal in the downlink global model transmission may be discarded from the training process in a particular iteration. Moreover, by leveraging error correction codes at the LEDs, will enable the clients to detect   certain errors in the corrupted global model, and then correct them to avoid global model re-transmission.

\textcolor{black}{ In traditional FL algorithms, the size of the training tasks and different training hyperparameters, such as the input dataset, batch size, epochs per round, and learning rates for different clients are specified at the beginning of the learning procedure and remain unchanged throughout the entire process. However, this negatively affects the accuracy of the global model and the convergence rate, due to  the heterogeneous clients' datasets, and their diverse computational and communication capabilities. Hence, speeding up the training process while achieving high accuracy level requires correct tuning of  these parameters. This can be accomplished by quantifying the capabilities of all clients, in order to assign appropriate portion of the task to each one. Then, by monitoring the training progress, different parameters can be tuned through model-based optimization techniques, taking into consideration the available computing power, memory, and bandwidth resources. Apart from computing, optimizing  the wireless network is important in improving the FL performance through overcoming VLC channel associated limitations, such as limited resources and  introduced errors, interference, and delays.}

\subsection{\textcolor{black}{Communication Efficiency}}
\begin{figure*}[t]
        \includegraphics
       [width=1\linewidth] {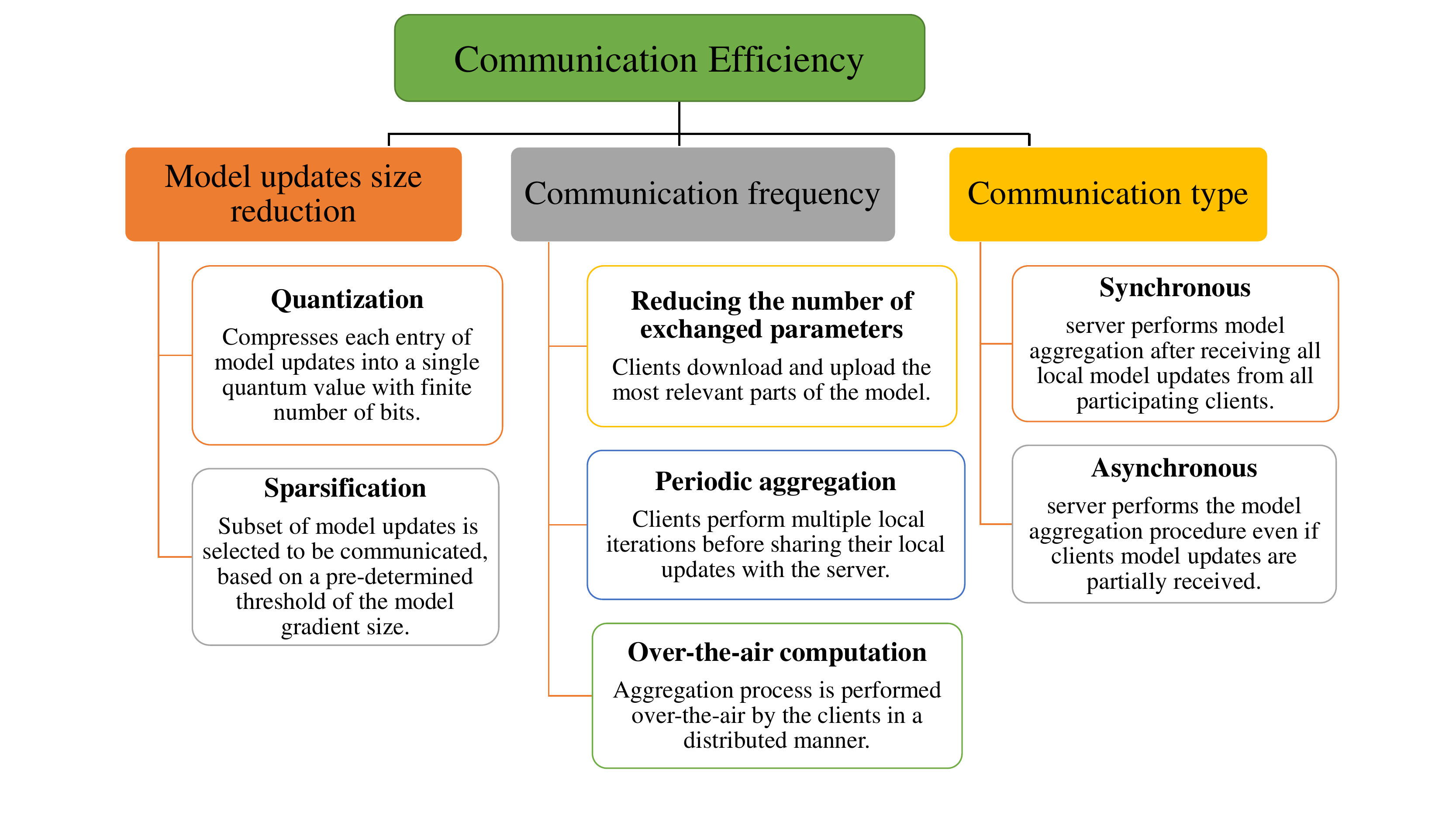}
    \caption{\textcolor{black}{Different communication efficiency enhancement schemes in FL.}  }
    \label{communication_efficiency}
\end{figure*}

It is recalled that  in large-scale FL-enabled networks, a large number of parameters updates need to be exchanged in each communication round. Hence,  research efforts have been devoted to achieve communication-efficient implementation of FL. Specifically,  three main directions exist, namely, model updates size reduction, communication frequency reduction, and communication type \cite{Wahab2021}. A detailed description of the three schemes and their variants is illustrated in Fig.\ref{communication_efficiency}.
%\subsection{Model Compression}

\subsection{Users Mobility Behaviour Prediction }
Modeling and predicting users  mobility in indoor VLC environment plays an important role in analyzing different communication design aspects. In particular, mobility prediction constitutes an efficient tool for location update, radio resource management, signaling traffic needed for handover, and users' association. In  FL, users mobility limits the performance of FL in VLC networks. This stems from the fact that the nature and amount of available training datasets vary with mobility, in addition to channel state information (CSI) fluctuation. 
Therefore,   enhancing   model training and   aggregation accuracy of local updates in FL, requires consideration of users mobility. 

Two different approaches are presented in the literature for individual mobility prediction, namely, personal mobility model with local-information and joint mobility model with population information. In the former, user's local mobility data is utilized to predict its own mobility behaviour. In such techniques, overcoming the sparsity of the mobility data records requires collaborative model training. To that end, FL can be leveraged, through a large number of clients,   to evaluate a global generalized model that can be utilized for mobility prediction.
Also, by leveraging mobility prediction models, clients selection can be preformed according to the mobility behavior of each user. Hence, only users with low mobility can participate in local model training, in order to prevent transmission errors that may occur during local model updates.

\section{Open research directions }

\subsection{Generative Adversarial Networks for Enhanced VLC Channel Estimation}

\begin{figure*}
     \centering
         \subfloat[]{ \includegraphics [width=.5\linewidth] {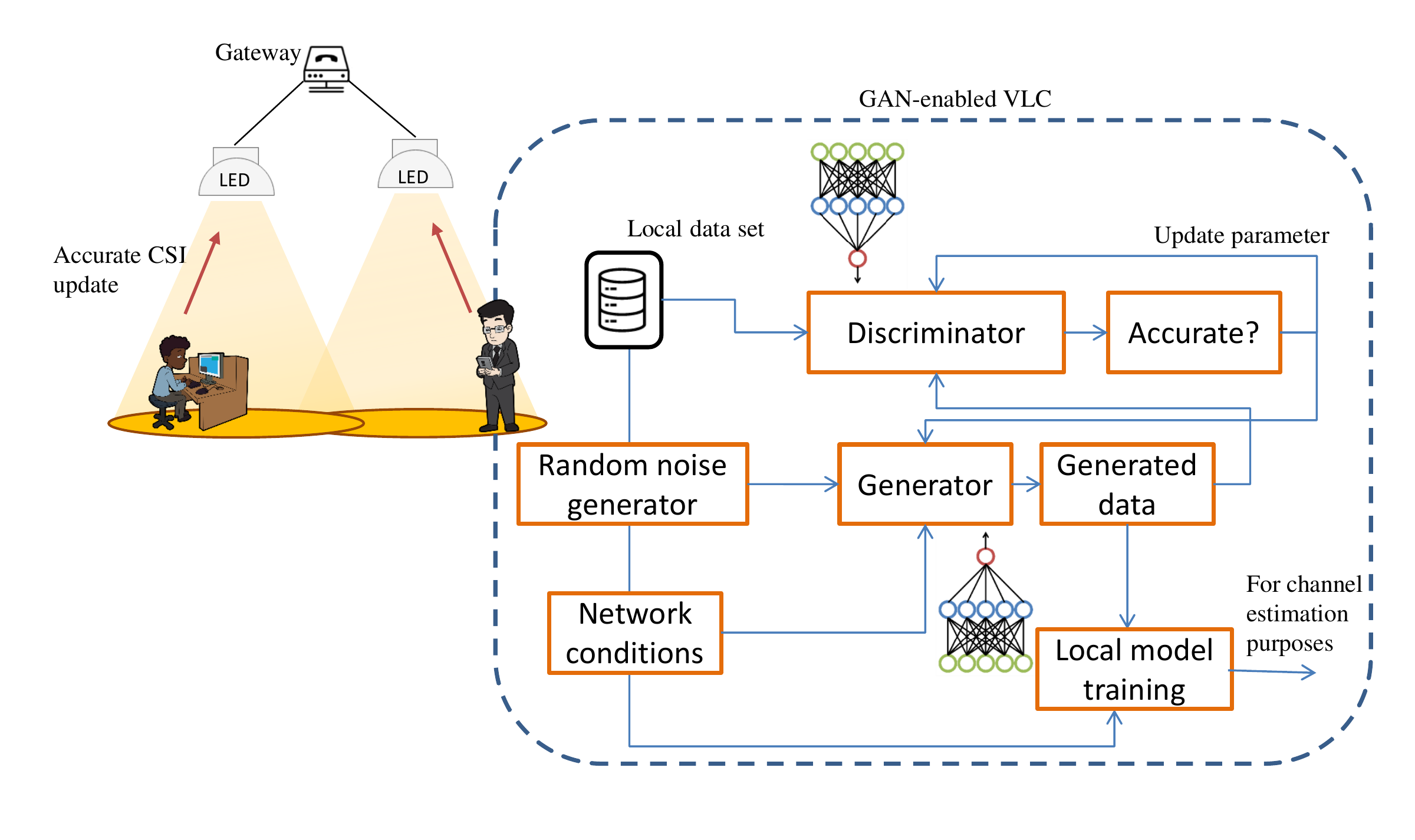}
 \label{Fig:GAN}}
\subfloat[]{ \includegraphics [width=.4\linewidth] {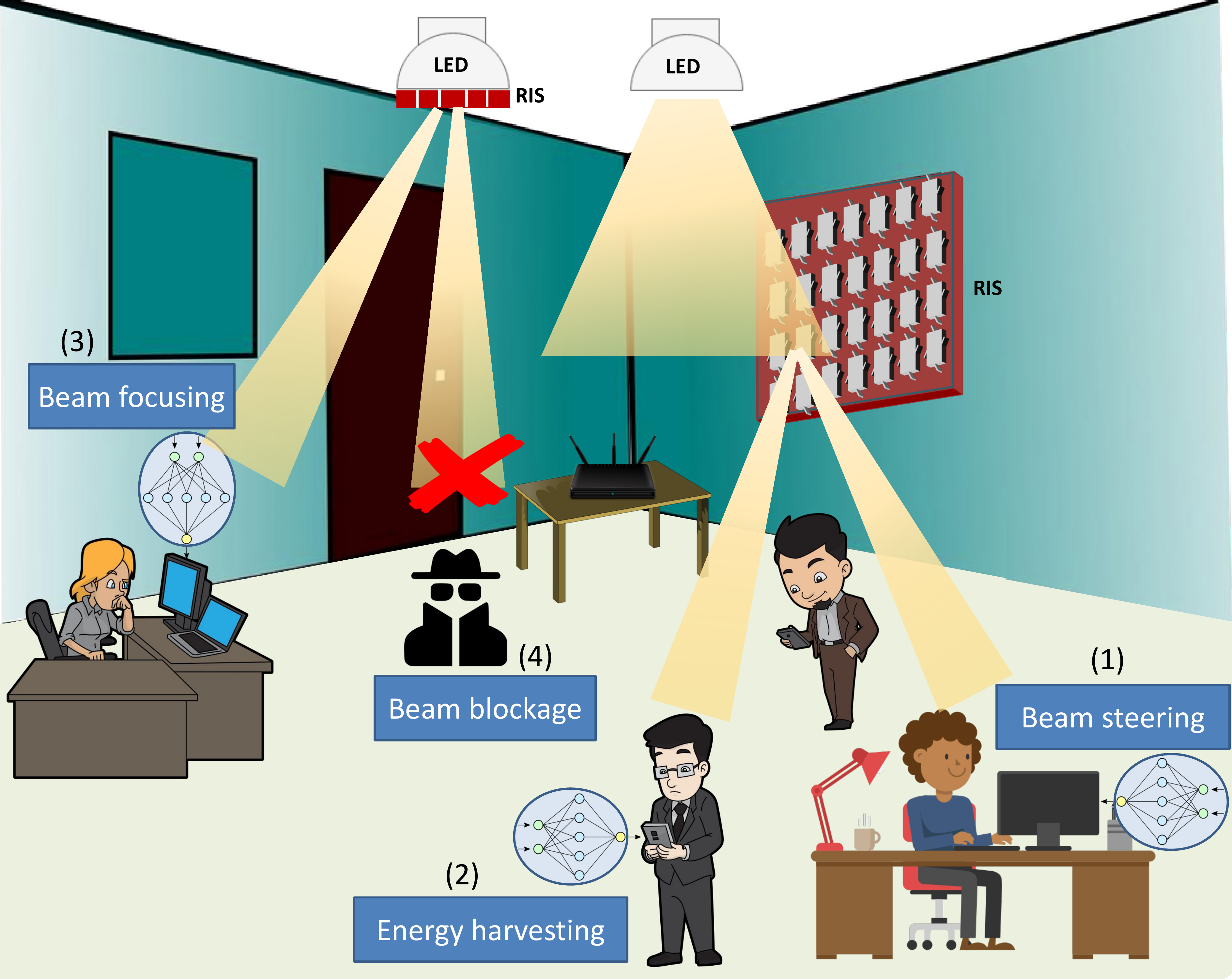}\label{Fig:RIS}}
     \hspace{0mm}
\subfloat[]{\includegraphics [width=.5\linewidth] {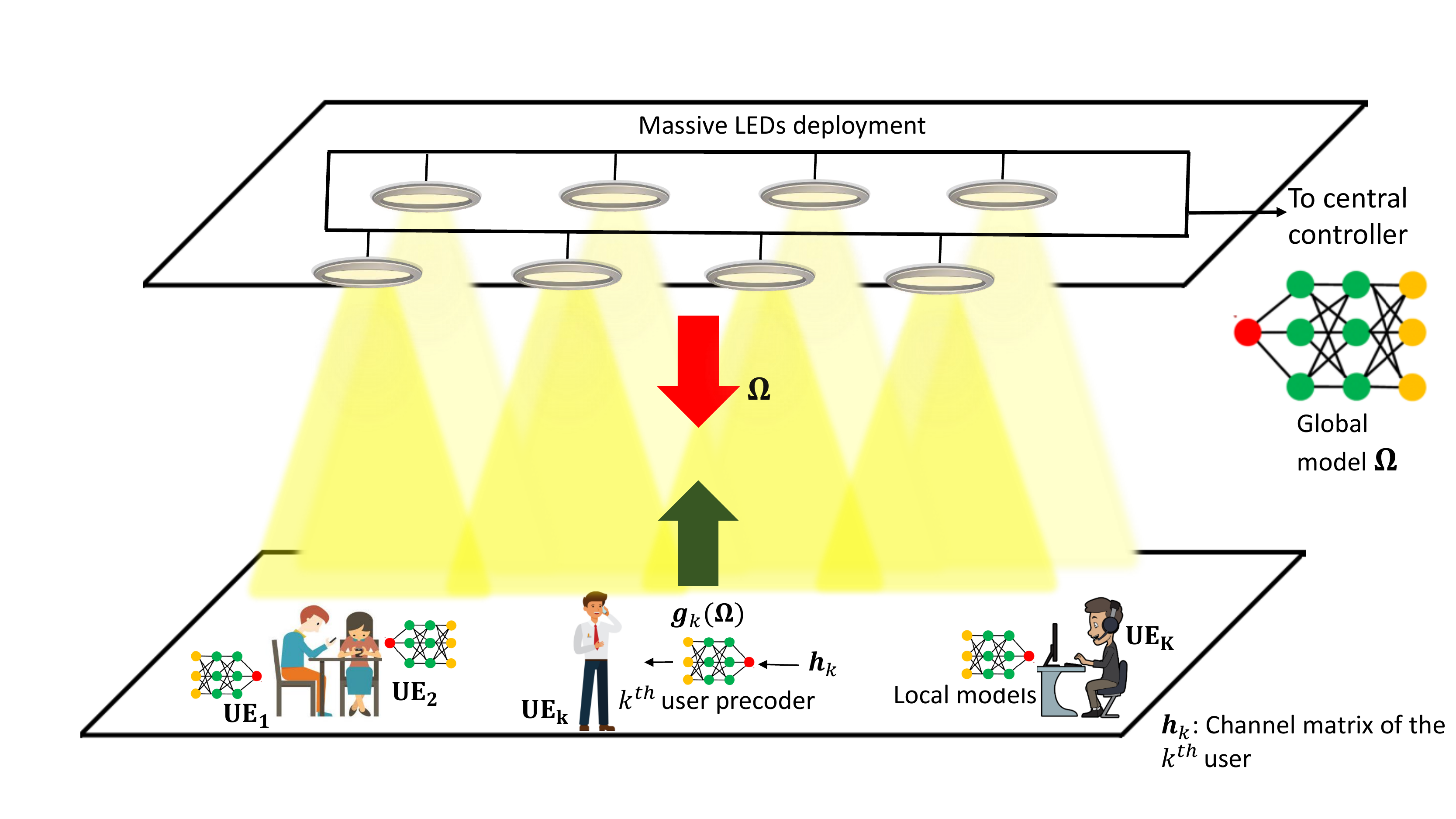}\label{Fig:MassiveMIMO}}
\label{Fig:GAN_RIS_MIMO}
\caption{(\ref{Fig:GAN}) GAN-enabled VLC system for enhanced CSI acquisition through FL. (\ref{Fig:RIS}) RIS-enabled FL for VLC scenario. (\ref{Fig:MassiveMIMO}) FL for Massive-MIMO VLC scenario.
%\label{Fig:futureworks}
}
    \end{figure*}

FL was recently considered    in  distributed CSI acquisition. Also, it was effective in data transmission overhead reduction  compared to centralized learning, while ensuring reliable model training and acceptable level of channel estimation accuracy \cite{elbir2020}.
In this context, local datasets at each participating device may fail to capture the VLC channels behavior in different scenarios, including the presence of ambient light noise, receiver random orientation, shadowing, and user mobility. In particular, when VLC channel conditions vary due to specific scenarios, it is essential to re-estimate the channel using pilots, and then collect data and update the local models accordingly. This will result in an increased pilot overhead, and therefore a higher loss in energy and time. In light of this, in order to develop generalized models for accurate channel estimation in VLC, local models should be trained while considering extreme network cases, imposing additional challenges on the implementation of FL in VLC networks.

To this end, generative adversarial networks (GANs) represent an efficient solution to create a generalized framework that experiences a wide range of special network conditions \cite{kasgari2020}. Specifically, in a GAN, a generator, which is enabled by a deep neural network, is trained to generate close-to-real channel data, and then a discriminator is utilized to quantify the learning accuracy. By leveraging GANs, the limited local datasets, representing the behavior of VLC channels under particular scenarios, will be extended to comprise real and synthetic data, covering all network conditions. Therefore,  improved  models training can be accomplished, and hence a more accurate and generalized channel estimation can be acquired. \textcolor{black}{ A typical GAN-enabled VLC system that utilizes FL in order to enhance CSI acquisition is depicted in  Fig. \ref{Fig:GAN}.} 
%From a different angle, GAN algorithm requires a large dataset for accurate samples generation. Therefore, the learning process, synthetic data generation accuracy, and hence, the reliability of the estimated channel behavior is constrained by the limited available channel samples at each device. In this regard, an extended version of GAN is proposed in \cite{zhang2021} to achieve network-wide channel estimation and modeling, in which all participating devices collaborate in order to develop generalized and comprehensive models. This developed GAN architecture is known as distributed GAN (DGAN). Unlike conventional distributed cooperative learning methods, where devices share their datasets for exhaustive training, DGAN brings up the advantage of ensuring users data privacy. In particular, participating devices in DGAN share locally generated synthetic data, instead of their private raw data. 
As a promising algorithm, research efforts should be directed towards implementing GANs in FL-based VLC systems, outlining implementation challenges, practical design aspects, and highlighting possible applications.  
\subsection{Reconfigurable Intelligence Surfaces}
The reconfigurable intelligent surface (RIS) concept was recently identified as a key enabler for beyond 5G networks,  offering extended coverage, enhanced signals reliability, and improved energy efficiency. RIS comprises a number of reconfigurable metasurfaces with unique artificially-manipulated electromagnetic properties, enabling them to control/adjust the properties of impinging wireless signals. This can be achieved by enabling a wide range of functionalities, including beam focusing, splitting, reflection, absorption, and polarization. Therefore, RIS can be of  particular interest in FL-enabled VLC systems from two different perspectives, namely, RIS-assisted and RIS-equipped LEDs. Regarding the former, multiple RISs can be mounted on the walls of an indoor area   to enable a number of functionalities that assist with global models transmission. Specifically, RIS can play a vital role in assisting the establishment of LoS links between participating devices and the server. It is recalled that having a LoS link is an essential component in VLC systems, and therefore, any blockage yields a service interruption. In this context, RIS can be a promising candidate to tackle this issue [Fig. \ref{Fig:RIS}, (1)]. Also,  signals reflected from the RIS can be used for energy harvesting purposes, allowing power-constrained device to communicate their models reliably [Fig. \ref{Fig:RIS}, (2)].

On the contrary, with a proper tuning of an RIS, which is placed at the transmitter front-end, beam focusing can be realized by a controlled adjustment of the LED's FoV. This results in an improved global model reception and increased number of participating devices attributed to the improved coverage [Fig. \ref{Fig:RIS}, (3)]. Within the same context, an RIS can be exploited  to enhance the physical layer security, by blocking the transmitted global models and prevent them from potential eavesdroppers [Fig. \ref{Fig:RIS}, (4)].
However, such promising advantages, attained by the integration of RIS in FL-enabled VLC systems, can be realized only if the RIS parameters are properly optimized and tuned to deliver the anticipated outcomes. It is worth highlighting that the optimization of FL-enabled VLC with RIS has not been touched in the literature yet, rendering it an attractive open research problem. 

\subsection{Multiple Access}
Multiple access (MA) schemes are indispensable parts of future network generations, fulfilling the massive scale connectivity associated with emerging applications. In VLC, several optical orthogonal and non-orthogonal multiple access schemes have been developed. 
%, including  time-division multiple access, orthogonal frequency-division multiple  access, and optical code-division  multiple  access. On the contrary, space-division multiple access exploits the spatial separation between users to provide full time and frequency resources. Furthermore, non-orthogonal multiple access has been recognized as a spectrally-efficient multiple access scheme that allows different users' signals to be multiplexed in power domain, sharing the same frequency resources simultaneously. 
Notably, these schemes fundamentally rely on advanced optimization algorithms, in order to coordinate users access to the network resources. Owing to the inherent non-convexity and infinite dimensionality of these optimization problems, iterative algorithms are usually exploited with the aim to obtain an optimum resource allocation, allowing fair users access to the network. Despite the satisfying performance achieved by different optimization tools in MA schemes, their performance is generally constrained by the high computational overhead, which hinders their real-time implementation. Moreover, due to the dynamic nature of VLC networks, a frequent execution of the iterative algorithms will occur.

Conventionally, classical ML constitutes the optimum tool to facilitate solving such optimization problems, with the aid of sensory data transmitted from the clients, such as current allocated spectrum, device non-linearity information, and the presence of interfering signals. However, centralized ML algorithms have shown some shortage in terms of privacy, delay, and energy consumption, and hence, FL can be a prominent alternative to generate locally trained models. In this regard, the global feedback mechanism in FL allows participating devices to utilize the globally trained model to perform on-site resource allocation optimization, and hence, achieve cooperative coordinated network access.  
 
\subsection{FL in Hybrid RF/VLC Systems}
Typically, in VLC, light emitted from LEDs is confined within small areas, limiting the participating devices to the ones exist in the LEDs coverage area. Since VLC can provide interference-free communication, with co-existing RF systems, hybrid integration of RF and VLC is expected to provide ubiquitous coverage and enhanced user experience. In a hybrid RF/VLC architecture, each LED serves as an AP to provide high data rate transmission, and is supported by one or multiple RF APs that guarantee uninterrupted moderate data rate transmission, in case of blockage. Hence, each user within the indoor environment is associated with either a VLC or RF AP.\\
\noindent It is recalled that user selection is one of the most challenging issues in FL, particularly in RF systems due to the limited resources. Therefore, hybrid RF/VLC architecture constitutes an appealing solution to enhance the performance of FL by: i) allowing a larger number of users to participate in the model training process; ii) establish communication links for VLC clients in case of blockage. However, to ensure efficient integration of FL in hybrid RF/VLC systems, its performance needs to be optimized by considering APs-users association and resource allocation.
\subsection{FL for Augmented Reality Applications in VLC }
Augmented reality (AR) application is one of the latest technology trends, emerged to provide interactive and immersive users experience, by combining virtual visual and auditory contents with real environments. AR spans a variety of applications in different disciplines starting from TV and films production, weather sciences, disaster relief, medicine, education, and entertainments. To provide immersive experience over the real world, these AR devices are equipped with cameras, \textcolor{black}{global positioning system} (GPS) modules, and sensors. However, AR applications are highly localized and sensitive to latency issues. Meanwhile, AR applications generate enormous data from multiple users such as images, which require intensive data processing capabilities and  bandwidth resources. Typical high quality AR applications require data rates of multiple Gbps. 
Moreover, with the accelerating demands for multi-object virtualization, the accuracy of detection and classification is essential to enhance users immersive experience. 
Hence, overcoming latency and enhancing clients privacy whilst reducing communication overhead,  demanding AR algorithms   can be processed at the AR users side with the aid of FL. Conversely, VLC is characterized by the ability to provide secure high data rate communication. Therefore, it can establish high speed wireless links between a centralized server and  AR clients to offload models update traffic from the current crowded RF spectrum, as a way to overcome the aforementioned limitations. Hence, integrating FL into VLC constitutes an important part in enhancing users experience in AR applications. 
\subsection{\textcolor{black}{FL and Massive MIMO VLC System}}
\textcolor{black}{VLC systems can benefit from the massive deployment of LEDs in order to enable massive multiple-input multiple-output (MIMO) configuration in a distributed manner. Massive-MIMO technology has shown a great potential to cater for the ever-increasing mobile data traffic through enhancing communication capacity and spectrum efficiency. Nevertheless, one of its major challenges is the need for efficient channel estimation techniques in order to obtain accurate CSI at the transmitter site, with the aim to design efficient precoders. In fact, channel estimation techniques require a huge  pilot overhead,  hindering the realization of massive-MIMO in VLC systems. Additionally, massive-MIMO VLC systems suffer from high inter-channel interference caused by the high spatial correlation between VLC channels. }

\textcolor{black}{Traditionally, precoder designs mainly rely on solving optimization problems in an iterative manner. However, the difficulty to obtain an optimum solution and the high computational complexity associated with these techniques are particularly challenging. 
%In this context, centralized ML techniques have shown a great success in handling complicated optimization problems for obtaining robust precoders design.  Usually, this is achieved by training a certain ML model at the transmitter side using the data collected from multiple users. Moreover, ML techniques are characterized by the ability to elicit new features from a limited-features training set. Nevertheless, as mentioned earlier, centralized ML algorithms come with their own challenges such as the increased transmission overhead, privacy, and propagation delay. Therefore,
Therefore, FL plays an important role in designing robust precoders for massive-MIMO VLC systems. To achieve this, each user is assumed to have its own training data pairs that consist of a channel matrix as an input and the precoder values as an output. During the training process, all the gradient values resulted from local training process at each user are aggregated at the central server. Once a targeted accuracy level is attained, the trained global model is shared with the users in order for each user to predict the corresponding precoder. It is worth mentioning that massive-MIMO VLC systems can also assist with enhancing the performance of FL, thanks to its high multiplexing gain that allows multiple FL tasks to be executed in parallel.  Fig. \ref{Fig:MassiveMIMO} illustrates a typical FL process utilized for efficient precoders designs in a multi-user massive-MIMO VLC system. }
\subsection{\textcolor{black}{Meta-Learning for Personalized FL in VLC}}
\textcolor{black}{
The non independent and identically distributed (iid) and personalized nature of local datasets of different clients represents one of the major challenges of traditional FL techniques. Such a challenge might be more pronounced in VLC systems, due to the inherent heterogeneity of local devices and their distinct activities and tasks. Thus, data tend to have different sizes, features, and target classes distribution. As a consequence, local models training over these heterogeneous datasets may result in inaccurate global model evaluation. In this regard,  meta-learning was recently promoted as a natural choice for federated setting, that is particularly well-suited for clients with statistically heterogeneous local datasets \cite{fallah2020}.  Meta-learning-based FL allows sharing a  parameterized algorithm, instead of a global model, enabling clients to learn the model parameters that can be adjusted very quickly according to the clients requirements, using only a few training examples. The application of meta-learning-based FL will allow participating clients to train a meta-model (algorithm parameters) in order to perform their own personalized tasks. Therefore, owning to their potentials in improving the performance of FL, the implementation of meta-learning techniques in FL-enabled VLC systems is a promising future research direction, that needs to be thoroughly investigated for the reliable implementation of FL in VLC environments.   }
\section{Conclusion}
We addressed the potentials of integrating the newly emerged FL paradigm in VLC systems. In particular, we presented a brief background about VLC technology and  the basic concepts of FL and aggregation mechanisms. Subsequently, we provided the fundamentals for  integrating FL into VLC systems and highlighted on its design aspects and  promising solutions. Finally, we outlined some envisioned future research directions, which need be investigated prior to real implementation of FL VLC systems.

\balance 
\bibliographystyle{IEEEtran}
\bibliography{IEEEabrv,FL_VLC_uncolored}

\end{document}